\documentclass[10pt,journal,cspaper,compsoc]{IEEEtran}

\usepackage{amsmath,amssymb}
\usepackage{algorithm}
\usepackage{algorithmic}
\usepackage{multirow}

\usepackage[aboveskip=8pt]{caption}

\usepackage[dvips]{graphicx}
\DeclareGraphicsExtensions{.pdf}
\usepackage[american]{babel}

\usepackage{tabularx}

\usepackage[bookmarks=false,colorlinks=true,linkcolor=black,citecolor=black,filecolor=black,urlcolor=black]{hyperref}
\usepackage{url}
\usepackage{cvpr}

\usepackage{multicol}

\newcommand{\cb}[1]{\textbf{#1}}

\newcommand{\ct}[1]{\fontsize{6pt}{1pt}\selectfont{#1}}
\newcolumntype{x}{>\small c}
\newcolumntype{y}{>\footnotesize c}

\begin{document}
\title{Spatial Pyramid Pooling in Deep Convolutional \\ Networks for Visual Recognition}

\author{Kaiming~He,
        Xiangyu~Zhang,
        Shaoqing~Ren,
        and~Jian~Sun
\IEEEcompsocitemizethanks{
\IEEEcompsocthanksitem K.~He and J.~Sun are with Microsoft Research, Beijing, China. E-mail: \{kahe,jiansun\}@microsoft.com
\IEEEcompsocthanksitem X. Zhang is with Xi'an Jiaotong University, Xi'an, China. Email: xyz.clx@stu.xjtu.edu.cn
\IEEEcompsocthanksitem S. Ren is with University of Science and Technology of China, Hefei, China. Email: sqren@mail.ustc.edu.cn}
\thanks{This work was done when X. Zhang and S. Ren were interns at Microsoft Research.}}

\markboth{}%
{He \MakeLowercase{\textit{et al.}}: Spatial Pyramid Pooling in Deep Convolutional Networks for Visual Recognition}

\IEEEcompsoctitleabstractindextext{%
\begin{abstract}
Existing deep convolutional neural networks (CNNs) require a fixed-size (\eg, 224$\times$224) input image. This requirement is ``artificial" and may reduce the recognition accuracy for the images or sub-images of an arbitrary size/scale. In this work, we equip the networks with another pooling strategy, ``spatial pyramid pooling", to eliminate the above requirement. The new network structure, called SPP-net, can generate a fixed-length representation regardless of image size/scale. Pyramid pooling is also robust to object deformations. With these advantages, SPP-net should in general improve all CNN-based image classification methods. On the ImageNet 2012 dataset, we demonstrate that SPP-net boosts the accuracy of a variety of CNN architectures despite their different designs. On the Pascal VOC 2007 and Caltech101 datasets, SPP-net achieves state-of-the-art classification results using a single full-image representation and no fine-tuning.

\quad The power of SPP-net is also significant in object detection. Using SPP-net, we compute the feature maps from the entire image only once, and then pool features in arbitrary regions (sub-images) to generate fixed-length representations for training the detectors. This method avoids repeatedly computing the convolutional features. In processing test images, our method is 24-102$\times$ faster than the R-CNN method, while achieving better or comparable accuracy on Pascal VOC 2007.

\quad In ImageNet Large Scale Visual Recognition Challenge (ILSVRC) 2014, our methods rank \#2 in object detection and \#3 in image classification among all 38 teams.
This manuscript also introduces the improvement made for this competition.
\end{abstract}

\begin{keywords}
Convolutional Neural Networks, Spatial Pyramid Pooling, Image Classification, Object Detection
\end{keywords}}

\maketitle

\section{Introduction}

We are witnessing a rapid, revolutionary change in our vision community, mainly caused by deep convolutional neural networks (CNNs) \cite{LeCun1989} and the availability of large scale training data~\cite{Deng2009}. Deep-networks-based approaches have recently been substantially improving upon the state of the art in image classification~\cite{Krizhevsky2012,Zeiler2013,Sermanet2013,Chatfield2014}, object detection~\cite{Girshick2014,Will2014,Sermanet2013}, many other recognition tasks~\cite{Razavian2014, Yaniv2014, Ning2014,Yunchao2014}, and even non-recognition tasks.

\begin{figure}[t]
\center
\includegraphics[width=1.0\linewidth]{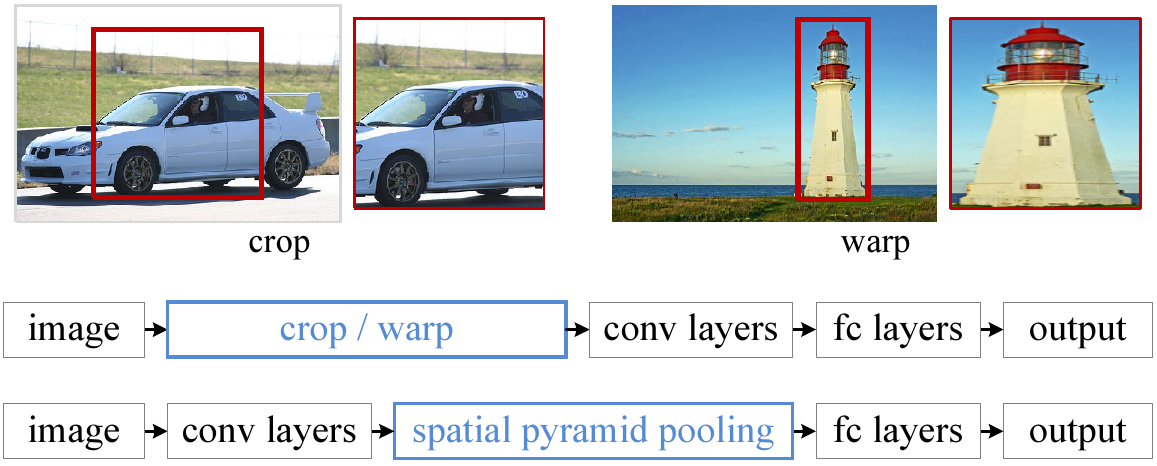}
\caption{Top: cropping or warping to fit a fixed size. Middle: a conventional CNN. Bottom: our spatial pyramid pooling network structure.}
\label{fig:teaser}
\end{figure}

However, there is a technical issue in the training and testing of the CNNs: the prevalent CNNs require a \emph{fixed} input image size (\eg, 224$\times$224), which limits both the aspect ratio and the scale of the input image. When applied to images of arbitrary sizes, current methods mostly fit the input image to the fixed size, either via cropping \cite{Krizhevsky2012,Zeiler2013} or via warping \cite{Donahue2013,Girshick2014}, as shown in Figure~\ref{fig:teaser} (top). But the cropped region may not contain the entire object, while the warped content may result in unwanted geometric distortion. Recognition accuracy can be compromised due to the content loss or distortion. Besides, a pre-defined scale may not be suitable when object scales vary. Fixing input sizes overlooks the issues involving scales.

\begin{figure*}[t]
\center
\includegraphics[width=1.0\linewidth]{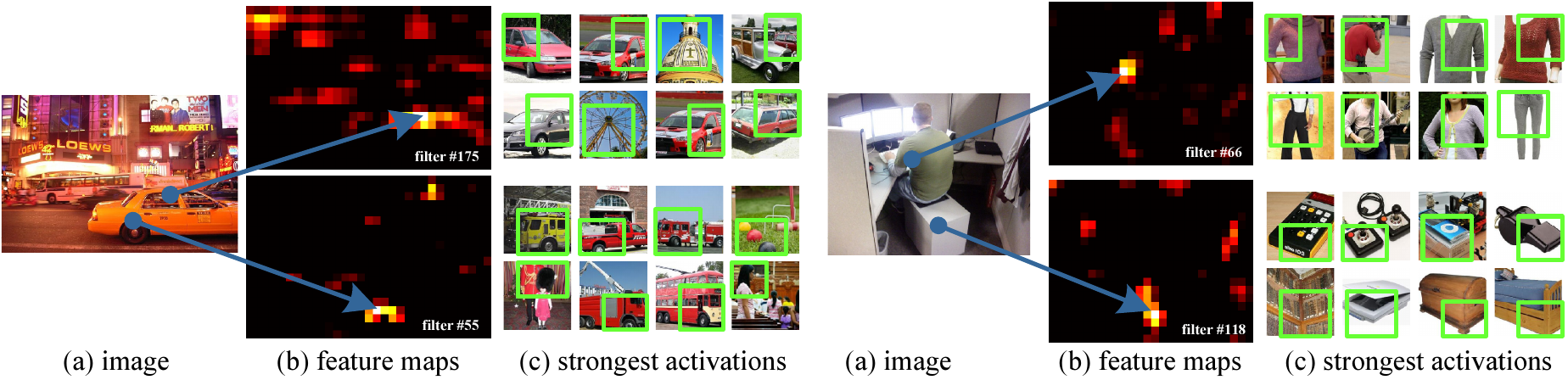}
\caption{Visualization of the feature maps. (a) Two images in Pascal VOC 2007. (b) The feature maps of some conv$_5$ filters. The arrows indicate the strongest responses and their corresponding positions in the images. (c) The ImageNet images that have the strongest responses of the corresponding filters. The green rectangles mark the receptive fields of the strongest responses.}
\label{fig:visual}
\end{figure*}

So why do CNNs require a fixed input size?
A CNN mainly consists of two parts: convolutional layers, and fully-connected layers that follow. The convolutional layers operate in a sliding-window manner and output feature maps which represent the spatial arrangement of the activations (Figure~\ref{fig:visual}). In fact, convolutional layers do not require a fixed image size and can generate feature maps of any sizes. On the other hand, the fully-connected layers need to have fixed-size/length input by their definition. Hence, the fixed-size constraint comes only from the fully-connected layers, which exist at a deeper stage of the network.

In this paper, we introduce a \emph{spatial pyramid pooling} (SPP) \cite{Grauman2005,Lazebnik2006} layer to remove the fixed-size constraint of the network. Specifically, we add an SPP layer on top of the last convolutional layer. The SPP layer pools the features and generates fixed-length outputs, which are then fed into the fully-connected layers (or other classifiers).
In other words, we perform some information ``aggregation" at a deeper stage of the network hierarchy (between convolutional layers and fully-connected layers) to avoid the need for cropping or warping at the beginning.
Figure~\ref{fig:teaser} (bottom) shows the change of the network architecture by introducing the SPP layer. We call the new network structure \emph{SPP-net}.


Spatial pyramid pooling \cite{Grauman2005,Lazebnik2006} (popularly known as spatial pyramid matching or SPM \cite{Lazebnik2006}), as an extension of the Bag-of-Words (BoW) model \cite{Sivic2003}, is one of the most successful methods in computer vision. It partitions the image into divisions from finer to coarser levels, and aggregates local features in them.
SPP has long been a key component in the leading and competition-winning systems for classification (\eg, \cite{Yang2009,Wang2010,Perronnin2010}) and detection (\eg, \cite{Sande2011}) before the recent prevalence of CNNs.
Nevertheless, SPP has not been considered in the context of CNNs. We note that SPP has several remarkable properties for deep CNNs: 1) SPP is able to generate a fixed-length output regardless of the input size, while the sliding window pooling used in the previous deep networks \cite{Krizhevsky2012} cannot; 2) SPP uses multi-level spatial bins, while the sliding window pooling uses only a single window size. Multi-level pooling has been shown to be robust to object deformations \cite{Lazebnik2006}; 3) SPP can pool features extracted at variable scales thanks to the flexibility of input scales.
Through experiments we show that all these factors elevate the recognition accuracy of deep networks.

SPP-net not only makes it possible to generate representations from arbitrarily sized images/windows for testing,
but also allows us to feed images with varying sizes or scales during training. Training with variable-size images increases scale-invariance and reduces over-fitting. We develop a simple multi-size training method. For a single network to accept variable input sizes, we approximate it by multiple networks that share all parameters, while each of these networks is trained using a fixed input size. In each epoch we train the network with a given input size, and switch to another input size for the next epoch. Experiments show that this multi-size training converges just as the traditional single-size training, and leads to better testing accuracy.

The advantages of SPP are orthogonal to the specific CNN designs. In a series of controlled experiments on the ImageNet 2012 dataset, we demonstrate that SPP improves four different CNN architectures in existing publications \cite{Krizhevsky2012,Zeiler2013,Sermanet2013} (or their modifications), over the no-SPP counterparts. These architectures have various filter numbers/sizes, strides, depths, or other designs.
It is thus reasonable for us to conjecture that SPP should improve more sophisticated (deeper and larger) convolutional architectures.
SPP-net also shows state-of-the-art classification results on Caltech101 \cite{Fei-Fei2007} and Pascal VOC 2007 \cite{Everingham2007} using only a \emph{single} full-image representation and no fine-tuning.

SPP-net also shows great strength in object detection.
In the leading object detection method R-CNN \cite{Girshick2014}, the features from candidate windows are extracted via deep convolutional networks. This method shows remarkable detection accuracy on both the VOC and ImageNet datasets. But the feature computation in R-CNN is time-consuming, because it repeatedly applies the deep convolutional networks to the raw pixels of thousands of warped regions per image. In this paper, we show that we can run the convolutional layers only \emph{once} on the entire image (regardless of the number of windows), and then extract features by SPP-net on the feature maps. This method yields a speedup of over one hundred times over R-CNN. Note that training/running a detector on the feature maps (rather than image regions) is actually a more popular idea \cite{Felzenszwalb2010,Dalal2005,Sande2011,Sermanet2013}.
But SPP-net inherits the power of the deep CNN feature maps and also the flexibility of SPP on arbitrary window sizes, which leads to outstanding accuracy and efficiency.
In our experiment, the SPP-net-based system (built upon the R-CNN pipeline) computes features 24-102$\times$ faster than R-CNN, while has better or comparable accuracy. With the recent fast proposal method of EdgeBoxes \cite{Zitnick2014}, our system takes 0.5 seconds processing an image (\emph{including all steps}). This makes our method practical for real-world applications.

A preliminary version of this manuscript has been published in ECCV 2014. Based on this work,
we attended the competition of ILSVRC 2014 \cite{Russakovsky2014},
and \textbf{ranked \#2 in object detection and \#3 in image classification} (both are provided-data-only tracks) among all 38 teams. There are a few modifications made for ILSVRC 2014. We show that the SPP-nets can boost various networks that are deeper and larger (Sec.~\ref{subsec:multi_level_pooling}-\ref{subsec:full_image}) over the no-SPP counterparts. Further, driven by our detection framework, we find that multi-view testing on feature maps with flexibly located/sized windows (Sec.~\ref{subsec:multi_view_testing}) can increase the classification accuracy. This manuscript also provides the details of these modifications.

We have released the code to facilitate future research (\emph{\small\url{http://research.microsoft.com/en-us/um/people/kahe/}}).

\section{Deep Networks with Spatial Pyramid Pooling}

\subsection{Convolutional Layers and Feature Maps}

Consider the popular seven-layer architectures \cite{Krizhevsky2012,Zeiler2013}. The first five layers are convolutional, some of which are followed by pooling layers. These pooling layers can also be considered as ``convolutional'', in the sense that they are using sliding windows. The last two layers are fully connected, with an N-way softmax as the output, where N is the number of categories.

The deep network described above needs a fixed image size.
However, we notice that the requirement of fixed sizes is only due to the fully-connected layers that demand fixed-length vectors as inputs. On the other hand, the convolutional layers accept inputs of arbitrary sizes. The convolutional layers use sliding filters, and their outputs have roughly the same aspect ratio as the inputs.
These outputs are known as \emph{feature maps} \cite{LeCun1989} - they involve not only the strength of the responses, but also their spatial positions.

In Figure~\ref{fig:visual}, we visualize some feature maps. They are generated by some filters of the conv$_5$ layer. Figure~\ref{fig:visual}(c) shows the strongest activated images of these filters in the ImageNet dataset. We see a filter can be activated by some semantic content. For example, the 55-th filter (Figure~\ref{fig:visual}, bottom left) is most activated by a circle shape; the 66-th filter (Figure~\ref{fig:visual}, top right) is most activated by a $\wedge$-shape; and the 118-th filter (Figure~\ref{fig:visual}, bottom right) is most activated by a $\vee$-shape. These shapes in the input images (Figure~\ref{fig:visual}(a)) activate the feature maps at the corresponding positions (the arrows in Figure~\ref{fig:visual}).

It is worth noticing that we generate the feature maps in Figure~\ref{fig:visual} without fixing the input size.
These feature maps generated by deep convolutional layers are analogous to the feature maps in traditional methods \cite{Chatfield2011,Coates2011}. In those methods, SIFT vectors \cite{Lowe2004} or image patches \cite{Coates2011} are densely extracted and then encoded, \eg, by vector quantization \cite{Sivic2003,Lazebnik2006,Gemert2008}, sparse coding \cite{Yang2009,Wang2010}, or Fisher kernels \cite{Perronnin2010}. These encoded features consist of the feature maps, and are then pooled by Bag-of-Words (BoW) \cite{Sivic2003} or spatial pyramids \cite{Grauman2005,Lazebnik2006}. Analogously, the deep convolutional features can be pooled in a similar way.

\begin{figure}[t]
\center
\includegraphics[width=1.0\linewidth]{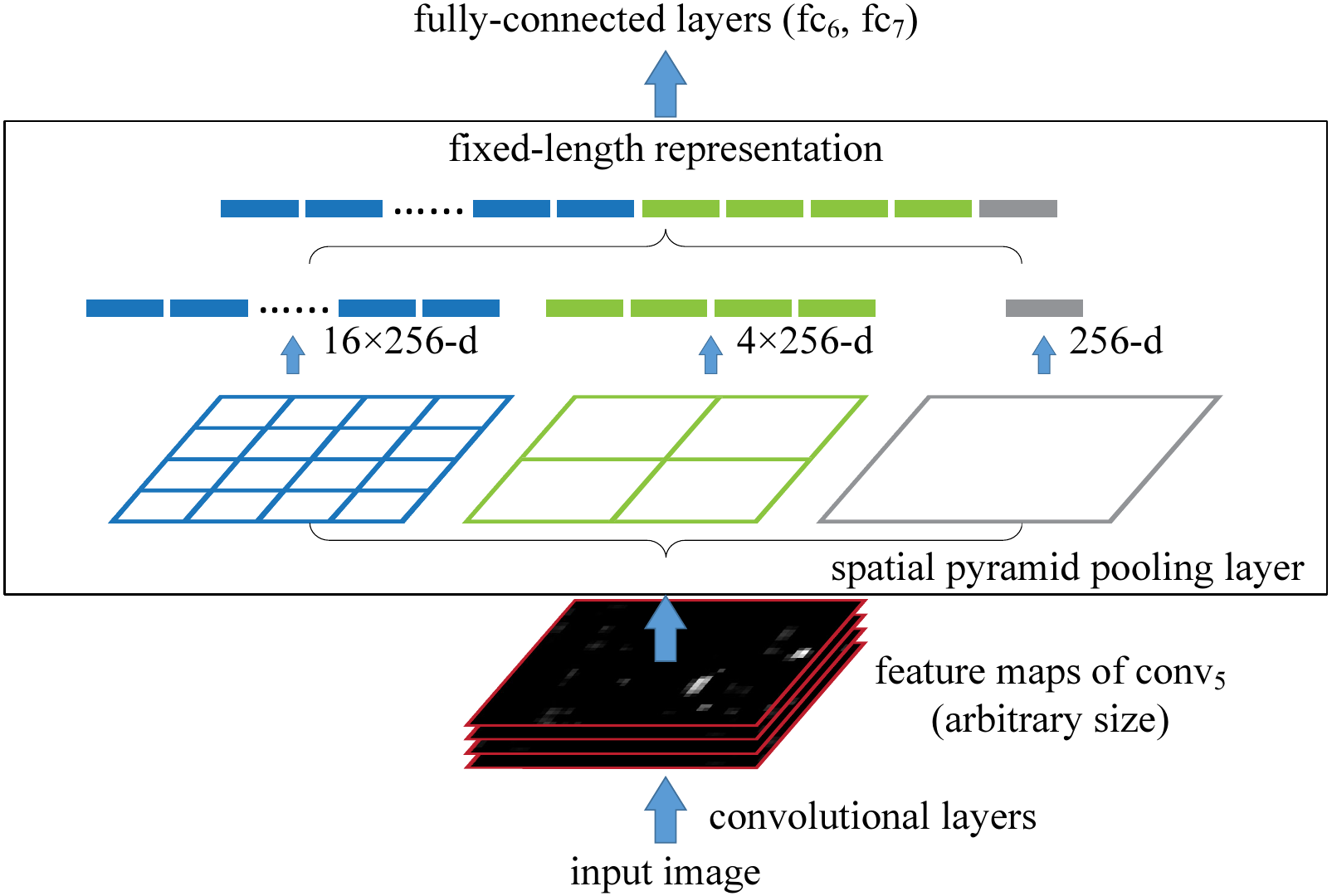}
\caption{A network structure with a \textbf{spatial pyramid pooling layer}. Here 256 is the filter number of the conv$_5$ layer, and conv$_5$ is the last convolutional layer.}
\label{fig:spm}
\end{figure}

\subsection{The Spatial Pyramid Pooling Layer}

The convolutional layers accept arbitrary input sizes, but they produce outputs of variable sizes. The classifiers (SVM/softmax) or fully-connected layers require fixed-length vectors. Such vectors can be generated by the Bag-of-Words (BoW) approach \cite{Sivic2003} that pools the features together. Spatial pyramid pooling \cite{Grauman2005,Lazebnik2006} improves BoW in that it can maintain spatial information by pooling in local spatial bins. These spatial bins have sizes proportional to the image size, so the number of bins is fixed regardless of the image size. This is in contrast to the sliding window pooling of the previous deep networks \cite{Krizhevsky2012}, where the number of sliding windows depends on the input size.

To adopt the deep network for images of arbitrary sizes, we replace the last pooling layer (\eg, pool$_5$, after the last convolutional layer) with a \emph{spatial pyramid pooling layer}. Figure~\ref{fig:spm} illustrates our method. In each spatial bin, we pool the responses of each filter (throughout this paper we use max pooling). The outputs of the spatial pyramid pooling are $kM$-dimensional vectors with the number of bins denoted as $M$ ($k$ is the number of filters in the last convolutional layer). The fixed-dimensional vectors are the input to the fully-connected layer.

With spatial pyramid pooling, the input image can be of any sizes. This not only allows arbitrary aspect ratios, but also allows arbitrary scales. We can resize the input image to any scale (\eg, $\min(w,h)$=180, 224, ...) and apply the same deep network. When the input image is at different scales, the network (with the same filter sizes) will extract features at different scales. The scales play important roles in traditional methods, \eg,
the SIFT vectors are often extracted at multiple scales \cite{Lowe2004,Chatfield2011} (determined by the sizes of the patches and Gaussian filters). We will show that the scales are also important for the accuracy of deep networks.

Interestingly, the coarsest pyramid level has a single bin that covers the entire image. This is in fact a ``global pooling'' operation, which is also investigated in several concurrent works. In \cite{Lin2013,Szegedy2014} a global average pooling is used to reduce the model size and also reduce overfitting; in \cite{Simonyan2014}, a global average pooling is used on the testing stage after all fc layers to improve accuracy; in \cite{Oquab2014}, a global max pooling is used for weakly supervised object recognition. The global pooling operation corresponds to the traditional Bag-of-Words method.

\subsection{Training the Network}

Theoretically, the above network structure can be trained with standard back-propagation \cite{LeCun1989}, regardless of the input image size. But in practice the GPU implementations (such as \emph{cuda-convnet} \cite{Krizhevsky2012} and \emph{Caffe} \cite{Jia2013}) are preferably run on fixed input images. Next we describe our training solution that takes advantage of these GPU implementations while still preserving the spatial pyramid pooling behaviors.

\subsubsection*{Single-size training}
\begin{figure}[t]
\center
\includegraphics[width=0.9\linewidth]{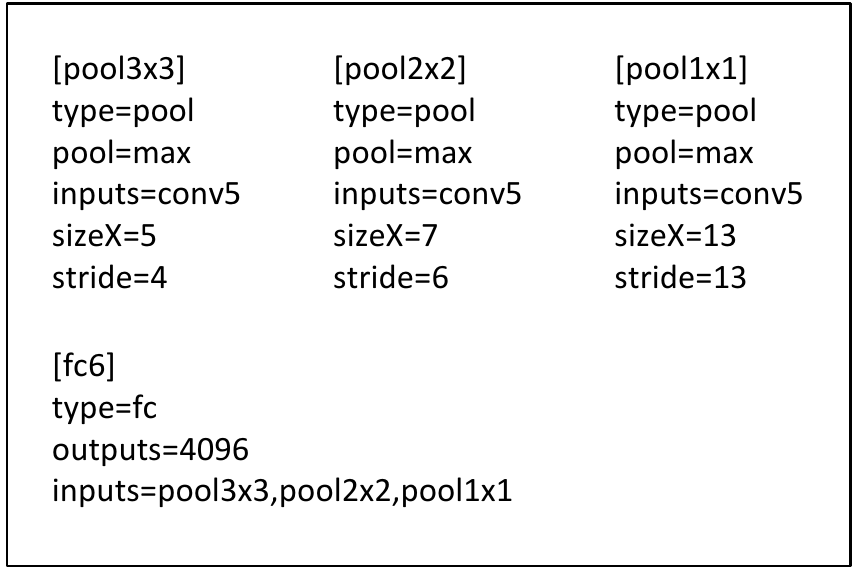}
\caption{An example 3-level pyramid pooling in the cuda-convnet style \cite{Krizhevsky2012}. Here sizeX is the size of the pooling window. This configuration is for a network whose feature map size of conv$_5$ is 13$\times$13, so the pool$_{3\times3}$, pool$_{2\times2}$, and pool$_{1\times1}$ layers will have 3$\times$3, 2$\times$2, and 1$\times$1 bins respectively.}
\label{fig:pseudo}
\end{figure}

As in previous works, we first consider a network taking a fixed-size input (224$\times$224) cropped from images. The cropping is for the purpose of data augmentation. For an image with a given size, we can pre-compute the bin sizes needed for spatial pyramid pooling. Consider the feature maps after conv$_5$ that have a size of $a$$\times$$a$ (\eg, 13$\times$13). With a pyramid level of $n$$\times$$n$ bins, we implement this pooling level as a sliding window pooling, where the window size $win=\lceil a/n \rceil$ and stride $str=\lfloor a/n \rfloor$ with $\lceil\cdot\rceil$ and $\lfloor\cdot\rfloor$ denoting ceiling and floor operations. With an $l$-level pyramid, we implement $l$ such layers. The next fully-connected layer (fc$_6$) will concatenate the $l$ outputs.
Figure~\ref{fig:pseudo} shows an example configuration of 3-level pyramid pooling (3$\times$3, 2$\times$2, 1$\times$1) in the \emph{cuda-convnet} style \cite{Krizhevsky2012}.

The main purpose of our single-size training is to enable the multi-level pooling behavior. Experiments show that this is one reason for the gain of accuracy.

\begin{table*}
\begin{center}
\begin{tabular}{c|c|c|c|c|c|c|c}
\hline
model & conv$_1$ & conv$_2$ & conv$_3$ & conv$_4$ & conv$_5$ & conv$_6$ & conv$_7$ \\
\hline
ZF-5 & $96\times7^2$, str~2 & $256\times5^2$, str~2 & $384\times3^2$ & $384\times3^2$ & $256\times3^2$ &  &  \\
          & LRN, pool $3^2$, str 2 & LRN, pool $3^2$, str 2 &  &  &  & - & - \\
   & map size $55\times55$ & $27\times27$ & $13\times13$ & $13\times13$ & $13\times13$ &  &  \\
\hline
Convnet*-5 & $96\times11^2$, str~4 & $256\times5^2$ & $384\times3^2$ & $384\times3^2$ & $256\times3^2$ &  &  \\
          & LRN, & LRN, pool $3^2$, str 2 & pool $3^2$, 2 &  &  & - & - \\
   & map size $55\times55$ & $27\times27$ & $13\times13$ & $13\times13$ & $13\times13$ &  &  \\
\hline
Overfeat-5/7 & $96\times7^2$, str~2 & $256\times5^2$ & $512\times3^2$ & $512\times3^2$ & $512\times3^2$ & $512\times3^2$ & $512\times3^2$\\
          & pool $3^2$, str 3, LRN & pool $2^2$, str 2 &  &  &  &  &  \\
   & map size $36\times36$ & $18\times18$ & $18\times18$ & $18\times18$ & $18\times18$ & $18\times18$ & $18\times18$\\
\hline
\end{tabular}
\end{center}
\caption{Network architectures: filter number$\times$filter size (\eg, $96\times7^2$), filter stride (\eg, str $2$), pooling window size (\eg, pool $3^2$), and the output feature map size (\eg, map size $55\times55$). LRN represents Local Response Normalization. The padding is adjusted to produce the expected output feature map size.}
\label{tab:architectures}
\end{table*}

\subsubsection*{Multi-size training}

Our network with SPP is expected to be applied on images of any sizes. To address the issue of varying image sizes in training, we consider a set of pre-defined sizes.
We consider two sizes: 180$\times$180 in addition to 224$\times$224. Rather than crop a smaller 180$\times$180 region, we resize the aforementioned 224$\times$224 region to 180$\times$180. So the regions at both scales differ only in resolution but not in content/layout. For the network to accept 180$\times$180 inputs, we implement another fixed-size-input (180$\times$180) network. The feature map size after conv$_5$ is $a$$\times$$a=10\times$10 in this case. Then we still use $win=\lceil a/n \rceil$ and $str=\lfloor a/n \rfloor$ to implement each pyramid pooling level.
The output of the spatial pyramid pooling layer of this 180-network has the same fixed length as the 224-network. As such, this 180-network has exactly the same parameters as the 224-network in each layer. In other words, during training we implement the varying-input-size SPP-net by two fixed-size networks that share parameters.

To reduce the overhead to switch from one network (\eg, 224) to the other (\eg, 180), we train each full epoch on one network, and then switch to the other one (keeping all weights) for the next full epoch. This is iterated.
In experiments, we find the convergence rate of this multi-size training to be similar to the above single-size training.

The main purpose of our multi-size training is to simulate the varying input sizes while still leveraging the existing well-optimized fixed-size implementations. Besides the above two-scale implementation, we have also tested a variant using $s\times s$ as input where $s$ is randomly and uniformly sampled from $[180, 224]$ at each epoch. We report the results of both variants in the experiment section.

Note that the above single/multi-size solutions are for training only. At the testing stage, it is straightforward to apply SPP-net on images of any sizes.

\section{SPP-net for Image Classification}
\label{sec:classification}

\subsection{Experiments on ImageNet 2012 Classification}

We train the networks on the 1000-category training set of ImageNet 2012. Our training algorithm follows the practices of previous work \cite{Krizhevsky2012,Zeiler2013,Howard2013}. The images are resized so that the smaller dimension is 256, and a 224$\times$224 crop is picked from the center or the four corners from the entire image\footnote{In \cite{Krizhevsky2012}, the four corners are picked from the corners of the central 256$\times$256 crop.}. The data are augmented by horizontal flipping and color altering \cite{Krizhevsky2012}. Dropout \cite{Krizhevsky2012} is used on the two fully-connected layers. The learning rate starts from 0.01, and is divided by 10 (twice) when the error plateaus. Our implementation is based on the publicly available code of \emph{cuda-convnet} \cite{Krizhevsky2012} and \emph{Caffe} \cite{Jia2013}. All networks in this paper can be trained on a single GeForce GTX Titan GPU (6 GB memory) within two to four weeks.

\newcommand{\bk}[1]{$_{\text{(#1)}}$}
\begin{table*}[t]
\small
\begin{center}
\begin{tabular}{cc|cccc}
\hline
 & & \multicolumn{4}{c}{top-1 error (\%)} \\
\hline
 & & ZF-5 & Convnet*-5 & Overfeat-5 & Overfeat-7\\
 \hline
 (a) & no SPP & 35.99 & 34.93 & 34.13 & 32.01\\
 (b) & SPP single-size trained  & 34.98 \bk{1.01} & 34.38 \bk{0.55} & 32.87 \bk{1.26} & 30.36 \bk{1.65}\\
 (c) & SPP multi-size trained &	34.60 \bk{1.39} & 33.94 \bk{0.99} & 32.26 \bk{1.87} & 29.68 \bk{2.33}\\
\hline
\end{tabular}\\
~\\
~\\
\begin{tabular}{cc|ccccc}
\hline
 & & \multicolumn{4}{c}{top-5 error (\%)} \\
\hline
 & & ZF-5 & Convnet*-5 & Overfeat-5 & Overfeat-7\\
 \hline
 (a) & no SPP & 14.76 & 13.92 & 13.52 & 11.97\\
(b) & SPP single-size trained & 14.14 \bk{0.62} & 13.54 \bk{0.38} & 12.80 \bk{0.72} & 11.12 \bk{0.85}\\
(c) & SPP multi-size trained & 13.64 \bk{1.12} & 13.33 \bk{0.59} & 12.33 \bk{1.19} & 10.95 \bk{1.02}\\
\hline
\end{tabular}
\end{center}
\caption{Error rates in the validation set of ImageNet 2012. All the results are obtained using standard 10-view testing. In the brackets are the gains over the ``no SPP'' baselines.}
\label{tab:imagenet0}
\end{table*}

\subsubsection{Baseline Network Architectures}

The advantages of SPP are independent of the convolutional network architectures used. We investigate four different network architectures in existing publications \cite{Krizhevsky2012,Zeiler2013,Sermanet2013} (or their modifications), and we show SPP improves the accuracy of all these architectures. These baseline architectures are in Table~\ref{tab:architectures} and briefly introduced below:

\begin{itemize}
  \item \textbf{ZF-5}: this architecture is based on Zeiler and Fergus's (ZF) ``fast'' (smaller) model \cite{Zeiler2013}. The number indicates five convolutional layers.
  \item \textbf{Convnet*-5}: this is a modification on Krizhevsky \etal's network \cite{Krizhevsky2012}. We put the two pooling layers after conv$_2$ and conv$_3$ (instead of after conv$_1$ and conv$_2$). As a result, the feature maps after each layer have the same size as ZF-5.
  \item \textbf{Overfeat-5/7}: this architecture is based on the Overfeat paper \cite{Sermanet2013}, with some modifications as in \cite{Chatfield2014}. In contrast to ZF-5/Convnet*-5, this architecture produces a larger feature map ($18\times18$ instead of $13\times13$) before the last pooling layer. A larger filter number (512) is used in conv$_3$ and the following convolutional layers. We also investigate a deeper architecture with 7 convolutional layers, where conv$_3$ to conv$_7$ have the same structures.
\end{itemize}
~\\
In the baseline models, the pooling layer after the last convolutional layer generates $6\times6$ feature maps, with two 4096-d fc layers and a 1000-way softmax layer following.
Our replications of these baseline networks are in Table~\ref{tab:imagenet0} (a). We train 70 epochs for ZF-5 and 90 epochs for the others.
Our replication of ZF-5 is better than the one reported in \cite{Zeiler2013}. This gain is because the corner crops are from the entire image, as is also reported in \cite{Howard2013}.

\subsubsection{Multi-level Pooling Improves Accuracy}
\label{subsec:multi_level_pooling}

In Table~\ref{tab:imagenet0} (b) we show the results using single-size training. The training and testing sizes are both 224$\times$224. In these networks, the convolutional layers have the same structures as the corresponding baseline models, whereas the pooling layer after the final convolutional layer is replaced with the SPP layer. For the results in Table~\ref{tab:imagenet0}, we use a 4-level pyramid. The pyramid is \{6$\times$6, 3$\times$3,  2$\times$2, 1$\times$1\} (totally 50 bins). For fair comparison, we still use the standard 10-view prediction with each view a 224$\times$224 crop.
Our results in Table~\ref{tab:imagenet0} (b) show considerable improvement over the no-SPP baselines in Table~\ref{tab:imagenet0} (a). Interestingly, the largest gain of top-1 error (1.65\%) is given by the most accurate architecture. Since we are still using the same 10 cropped views as in (a), these gains are solely because of multi-level pooling.

It is worth noticing that the gain of multi-level pooling is \textbf{not} simply due to more parameters; rather, it is because the multi-level pooling is robust to the variance in object deformations and spatial layout \cite{Lazebnik2006}. To show this, we train another ZF-5 network with a different 4-level pyramid: \{4$\times$4, 3$\times$3,  2$\times$2, 1$\times$1\} (totally 30 bins). This network has fewer parameters than its no-SPP counterpart, because its fc$_6$ layer has 30$\times$256-d inputs instead of 36$\times$256-d. The top-1/top-5 errors of this network are 35.06/14.04. This result is similar to the 50-bin pyramid above (34.98/14.14), but considerably better than the no-SPP counterpart (35.99/14.76).

\setlength{\tabcolsep}{3pt}
\begin{table}[t]
\footnotesize
\begin{center}
\begin{tabular}{c|c|c}
\hline
 SPP on & test view & top-1 val \\
\hline
ZF-5, single-size trained & 1 crop & 38.01\\
ZF-5, single-size trained & 1 full & \textbf{37.55}\\
\hline
ZF-5, multi-size trained & 1 crop & 37.57\\
ZF-5, multi-size trained & 1 full & \textbf{37.07}\\
\hline
\hline
Overfeat-7, single-size trained & 1 crop & 33.18\\
Overfeat-7, single-size trained & 1 full & \textbf{32.72}\\
\hline
Overfeat-7, multi-size trained & 1 crop & 32.57\\
Overfeat-7, multi-size trained & 1 full & \textbf{31.25}\\
\hline
\end{tabular}
\end{center}
\caption{Error rates in the validation set of ImageNet 2012 using a single view. The images are resized so $\min(w,h)=256$. The crop view is the central 224$\times$224 of the image.}
\label{tab:imagenet_fullview}
\end{table}

\subsubsection{Multi-size Training Improves Accuracy}

Table~\ref{tab:imagenet0} (c) shows our results using multi-size training. The training sizes are 224 and 180, while the testing size is still 224. We still use the standard 10-view prediction. The top-1/top-5 errors of all architectures further drop. The top-1 error of SPP-net (Overfeat-7) drops to 29.68\%, which is 2.33\% better than its no-SPP counterpart and 0.68\% better than its single-size trained counterpart.

Besides using the two discrete sizes of 180 and 224, we have also evaluated using a random size uniformly sampled from $[180, 224]$. The top-1/5 error of SPP-net (Overfeat-7) is 30.06\%/10.96\%. The top-1 error is slightly worse than the two-size version, possibly because the size of 224 (which is used for testing) is visited less.
But the results are still better the single-size version.

There are previous CNN solutions \cite{Sermanet2013,Howard2013} that deal with various scales/sizes,
but they are mostly based on testing.
In Overfeat \cite{Sermanet2013} and Howard's method \cite{Howard2013}, the single network is applied at multiple scales in the testing stage, and the scores are averaged. Howard further trains two different networks on low/high-resolution image regions and averages the scores. To our knowledge, our method is the first one that \emph{trains} a single network with input images of multiple sizes.

\setlength{\tabcolsep}{6pt}
\begin{table*}[t]
\small
\begin{center}
\begin{tabular}{c|c|c|ccc}
\hline
 method & test scales & test views & top-1 val & top-5 val & \textbf{top-5 test}\\
\hline
Krizhevsky \etal \cite{Krizhevsky2012} & 1 & 10 &40.7 & 18.2\\
\hline
Overfeat (fast) \cite{Sermanet2013} & 1 & - & 39.01 & 16.97\\
Overfeat (fast) \cite{Sermanet2013} & 6 & - &38.12 & 16.27\\
Overfeat (big) \cite{Sermanet2013} & 4 & - & 35.74 & 14.18\\
\hline
Howard (base) \cite{Howard2013} & 3 & 162 & 37.0 & 15.8\\
Howard (high-res) \cite{Howard2013} & 3 & 162 & 36.8 & 16.2\\
\hline
Zeiler \& Fergus (ZF) (fast) \cite{Zeiler2013} & 1 & 10 & 38.4 & 16.5\\
Zeiler \& Fergus (ZF) (big) \cite{Zeiler2013} & 1 & 10 & 37.5 & 16.0\\
\hline
Chatfield \etal \cite{Chatfield2014} & 1 & 10 & - & 13.1\\
\hline
ours (SPP O-7) & 1 & 10 & 29.68 & 10.95\\
ours (SPP O-7) & 6 & 96+2full & \textbf{27.86} & \textbf{9.14} & \textbf{9.08}\\
\hline
\end{tabular}
\end{center}
\caption{Error rates in ImageNet 2012. All the results are based on \textbf{a single network}. The number of views in Overfeat depends on the scales and strides, for which there are several hundreds at the finest scale.}
\label{tab:imagenet}
\end{table*}

\subsubsection{Full-image Representations Improve Accuracy}
\label{subsec:full_image}

Next we investigate the accuracy of the full-image views. We resize the image so that $\min(w,h)$=256 while maintaining its aspect ratio. The SPP-net is applied on this full image to compute the scores of the full view. For fair comparison, we also evaluate the accuracy of the single view in the center 224$\times$224 crop (which is used in the above evaluations). The comparisons of single-view testing accuracy are in Table~\ref{tab:imagenet_fullview}. Here we evaluate ZF-5/Overfeat-7.
The top-1 error rates are all reduced by the full-view representation. This shows the importance of maintaining the complete content. Even though our network is trained using square images only, it generalizes well to other aspect ratios.

Comparing Table~\ref{tab:imagenet0} and Table~\ref{tab:imagenet_fullview}, we find that the combination of multiple views is substantially better than the single full-image view. However, the full-image representations are still of good merits. First, we empirically find that (discussed in the next subsection) even for the combination of dozens of views, the additional two full-image views (with flipping) can still boost the accuracy by about 0.2\%. Second, the full-image view is methodologically consistent with the traditional methods \cite{Lazebnik2006,Yang2009,Perronnin2010} where the encoded SIFT vectors of the entire image are pooled together. Third, in other applications such as image retrieval \cite{Jegou2012}, an image representation, rather than a classification score, is required for similarity ranking. A full-image representation can be preferred.

\subsubsection{Multi-view Testing on Feature Maps}
\label{subsec:multi_view_testing}

Inspired by our detection algorithm (described in the next section), we further propose a multi-view testing method on the feature maps. Thanks to the flexibility of SPP, we can easily extract the features from windows (views) of arbitrary sizes from the convolutional feature maps.

On the testing stage, we resize an image so $\min(w,h)=s$ where $s$ represents a predefined scale (like 256). Then we compute the convolutional feature maps from the entire image. For the usage of flipped views, we also compute the feature maps of the flipped image.
Given any view (window) in the image, we map this window to the feature maps (the way of mapping is in Appendix), and then use SPP to pool the features from this window (see Figure~\ref{fig:spm_det}). The pooled features are then fed into the fc layers to compute the softmax score of this window. These scores are averaged for the final prediction. For the standard 10-view, we use $s=256$ and the views are 224$\times$224 windows on the corners or center. Experiments show that the top-5 error of the 10-view prediction on feature maps is within 0.1\% around the original 10-view prediction on image crops.

We further apply this method to extract multiple views from multiple scales. We resize the image to six scales $s\in\{224,256,300,360,448,560\}$ and compute the feature maps on the entire image for each scale. We use $224\times224$ as the view size for any scale, so these views have different relative sizes on the original image for different scales. We use 18 views for each scale: one at the center, four at the corners, and four on the middle of each side, with/without flipping (when $s$ = 224 there are 6 different views). The combination of these 96 views reduces the top-5 error from 10.95\% to 9.36\%. Combining the two full-image views (with flipping) further reduces the top-5 error to 9.14\%.

In the Overfeat paper \cite{Sermanet2013}, the views are also extracted from the convolutional feature maps instead of image crops. However, their views cannot have arbitrary sizes; rather, the windows are those where the pooled features match the desired dimensionality. We empirically find that these restricted windows are less beneficial than our flexibly located/sized windows.

\setlength{\tabcolsep}{8pt}
\begin{table}[t]
\small
\begin{center}
\begin{tabular}{cc|c}
\hline
rank & team & top-5 test\\
\hline
1 &  GoogLeNet \cite{Szegedy2014} & \textbf{6.66}\\
2 & VGG \cite{Simonyan2014} & 7.32\\
3 & \underline{ours} & \underline{8.06}\\
4 & Howard & 8.11\\
5 & DeeperVision & 9.50\\
6 & NUS-BST & 9.79\\
7 & TTIC\_ECP & 10.22\\
\hline
\end{tabular}
\caption{The competition results of ILSVRC 2014 classification \cite{Russakovsky2014}. The best entry of each team is listed.}
\label{tab:ilsvrc14_CLS}
\end{center}
\end{table}

\setlength{\tabcolsep}{4pt}
\begin{table*}[t]
\footnotesize
\begin{center}
\begin{tabular}{c|cccccc}
\hline
      & (a)       & (b) & (c) & (d) & (e)\\
model & no SPP (ZF-5) & SPP (ZF-5) & SPP (ZF-5) & SPP (ZF-5) & SPP (Overfeat-7)\\
\hline
     &                    crop     &    crop    &  full & full & full\\
size &                224$\times$224 & 224$\times$224 & 224$\times$- & 392$\times$- & 364$\times$- \\
\hline
conv$_4$ &                59.96 & 57.28 & -  & - & - \\
conv$_5$ &                66.34 & 65.43 & -  & - & - \\
pool$_{5/7}$ (6$\times$6) &   69.14 & 68.76 & 70.82 & 71.67 & 76.09 \\
fc$_{6/8}$ &                  74.86 & 75.55 & 77.32 & 78.78 & 81.58 \\
fc$_{7/9}$ &                  \underline{75.90} & \underline{76.45} & \underline{78.39} & \underline{80.10} & \underline{\textbf{82.44}} \\
\hline
\end{tabular}
\end{center}
\caption{Classification mAP in Pascal VOC 2007. For SPP-net, the pool$_{5/7}$ layer uses the 6$\times$6 pyramid level.}
\label{tab:voc2007}
\end{table*}

\setlength{\tabcolsep}{8pt}
\begin{table*}[t]
\footnotesize
\begin{center}
\begin{tabular}{c|c c c c}
\hline
      & (a)       & (b) & (c) & (d)\\
model & no SPP (ZF-5) & SPP (ZF-5) & SPP (ZF-5) & SPP (Overfeat-7)\\
\hline
     &                    crop     &    crop    &  full &  full\\
size &                224$\times$224 & 224$\times$224 & 224$\times$- & 224$\times$-\\
\hline
conv$_4$ &              80.12 & 81.03 & -  & - \\
conv$_5$ &              84.40 & 83.76 & - & - \\
pool$_{5/7}$ (6$\times$6) & \underline{87.98} & 87.60 & 89.46 & 91.46\\
SPP pool$_{5/7}$          &   -   & \underline{89.47} & \underline{91.44} &  \underline{\textbf{93.42}} \\
fc$_{6/8}$ &                87.86 & 88.54 & 89.50 & 91.83 \\
fc$_{7/9}$ &                85.30 & 86.10 & 87.08 & 90.00 \\
\hline
\end{tabular}
\caption{Classification accuracy in Caltech101. For SPP-net, the pool$_{5/7}$ layer uses the 6$\times$6 pyramid level.}
\label{tab:caltech101}
\end{center}
\end{table*}

\subsubsection{Summary and Results for ILSVRC 2014}

In Table~\ref{tab:imagenet} we compare with previous state-of-the-art methods. Krizhevsky \etal's \cite{Krizhevsky2012} is the winning method in ILSVRC 2012; Overfeat \cite{Sermanet2013}, Howard's \cite{Howard2013}, and Zeiler and Fergus's \cite{Zeiler2013} are the leading methods in ILSVRC 2013. We only consider single-network performance for manageable comparisons.

Our best single network achieves \textbf{9.14\%} top-5 error on the validation set. This is exactly the single-model entry we submitted to ILSVRC 2014 \cite{Russakovsky2014}. The top-5 error is \textbf{9.08\%} on the testing set (ILSVRC 2014 has the same training/validation/testing data as ILSVRC 2012). 
After combining eleven models, our team's result (\textbf{8.06}\%) is ranked \#3 among all 38 teams attending ILSVRC 2014 (Table~\ref{tab:ilsvrc14_CLS}).
Since the advantages of SPP-net should be in general independent of architectures, we expect that it will further improve the deeper and larger convolutional architectures \cite{Simonyan2014,Szegedy2014}.

\subsection{Experiments on VOC 2007 Classification}

Our method can generate a full-view image representation.
With the above networks pre-trained on ImageNet, we extract these representations from the images in the target datasets and re-train SVM classifiers \cite{Chang2011}. In the SVM training, we intentionally do not use any data augmentation (flip/multi-view). We l$_2$-normalize the features for SVM training.

The classification task in Pascal VOC 2007 \cite{Everingham2007} involves 9,963 images in 20 categories. 5,011 images are for training, and the rest are for testing. The performance is evaluated by mean Average Precision (mAP). Table~\ref{tab:voc2007} summarizes the results.

We start from a baseline in Table~\ref{tab:voc2007}~(a). The model is ZF-5 without SPP. To apply this model, we resize the image so that its smaller dimension is 224, and crop the center 224$\times$224 region.
The SVM is trained via the features of a layer. On this dataset, the deeper the layer is, the better the result is. In Table~\ref{tab:voc2007}~(b), we replace the no-SPP net with our SPP-net. As a first-step comparison, we still apply the SPP-net on the center 224$\times$224 crop. The results of the fc layers improve. This gain is mainly due to multi-level pooling.

Table~\ref{tab:voc2007} (c) shows our results on full images, where the images are resized so that the shorter side is 224. We find that the results are considerably improved (78.39\% \vs 76.45\%). This is due to the full-image representation that maintains the complete content.

Because the usage of our network does not depend on scale, we resize the images so that the smaller dimension is $s$ and use the same network to extract features. We find that $s=392$ gives the best results (Table~\ref{tab:voc2007} (d)) based on the validation set.
This is mainly because the objects occupy smaller regions in VOC 2007 but larger regions in ImageNet, so the relative object scales are different between the two sets. These results indicate scale matters in the classification tasks, and SPP-net can partially address this ``scale mismatch" issue.

In Table~\ref{tab:voc2007} (e) the network architecture is replaced with our best model (Overfeat-7, multi-size trained), and the mAP increases to \textbf{82.44\%}.
Table~\ref{tab:twosets} summarizes our results and the comparisons with the state-of-the-art methods. Among these methods, VQ \cite{Lazebnik2006}, LCC \cite{Wang2010}, and FK \cite{Perronnin2010} are all based on spatial pyramids matching, and \cite{Donahue2013,Zeiler2013,Oquab2014,Chatfield2014} are based on deep networks. In these results, Oquab \etal's (77.7\%) and Chatfield \etal's (82.42\%) are obtained by network fine-tuning and multi-view testing. Our result is comparable with the state of the art, using only a single full-image representation and without fine-tuning.

\subsection{Experiments on Caltech101}

The Caltech101 dataset \cite{Fei-Fei2007} contains 9,144 images in 102 categories (one background). We randomly sample 30 images per category for training and up to 50 images per category for testing. We repeat 10 random splits and average the accuracy. Table~\ref{tab:caltech101} summarizes our results.

There are some common observations in the Pascal VOC 2007 and Caltech101 results: SPP-net is better than the no-SPP net (Table~\ref{tab:caltech101} (b) \vs (a)), and the full-view representation is better than the crop ((c) \vs (b)). But the results in Caltech101 have some differences with Pascal VOC. The fully-connected layers are less accurate, and the SPP layers are better. This is possibly because the object categories in Caltech101 are less related to those in ImageNet, and the deeper layers are more category-specialized.
Further, we find that the scale 224 has the best performance among the scales we tested on this dataset. This is mainly because the objects in Caltech101 also occupy large regions of the images, as is the case of ImageNet.

Besides cropping, we also evaluate warping the image to fit the 224$\times$224 size. This solution maintains the complete content, but introduces distortion. On the SPP (ZF-5) model, the accuracy is 89.91\% using the SPP layer as features - lower than 91.44\% which uses the same model on the undistorted full image.

Table~\ref{tab:twosets} summarizes our results compared with the state-of-the-art methods on Caltech101. Our result (\textbf{93.42\%}) exceeds the previous record (88.54\%) by a substantial margin (4.88\%).

\begin{table}[t]
\small
\begin{center}
\begin{tabular}{c|c c}
\hline
   method        & VOC 2007 & Caltech101\\
\hline
\hline
VQ \cite{Lazebnik2006}$^\dag$ & 56.07 & 74.41$\pm$1.0\\
LLC \cite{Wang2010}$^\dag$ & 57.66 & 76.95$\pm$0.4\\
FK \cite{Perronnin2010}$^\dag$ & 61.69 & 77.78$\pm$0.6 \\
\hline
DeCAF \cite{Donahue2013} & - & 86.91$\pm$0.7 \\
Zeiler \& Fergus \cite{Zeiler2013} & 75.90$^\ddag$ & 86.5$\pm$0.5\\
Oquab \etal \cite{Oquab2014} & 77.7 & -\\
Chatfield \etal \cite{Chatfield2014} & \textbf{82.42} & 88.54$\pm$0.3 \\
\hline
ours & \textbf{82.44} & \textbf{93.42$\pm$0.5}\\
\hline
\end{tabular}
\end{center}
\caption{Classification results for Pascal VOC 2007 (mAP) and Caltech101 (accuracy). $^\dag$numbers reported by \cite{Chatfield2011}. $^\ddag$our implementation as in Table~\ref{tab:voc2007} (a).}
\label{tab:twosets}
\end{table}

\section{SPP-net for Object Detection}
\label{sec:detection}

Deep networks have been used for object detection. We briefly review the recent state-of-the-art R-CNN method \cite{Girshick2014}.
R-CNN first extracts about 2,000 candidate windows from each image via selective search \cite{Sande2011}. Then the image region in each window is warped to a fixed size (227$\times$227). A pre-trained deep network is used to extract the feature of each window. A binary SVM classifier is then trained on these features for detection. R-CNN generates results of compelling quality and substantially outperforms previous methods. However, because R-CNN repeatedly applies the deep convolutional network to about 2,000 windows per image, it is time-consuming. Feature extraction is the major timing bottleneck in testing.

\begin{figure}[t]
\center
\includegraphics[width=1.0\linewidth]{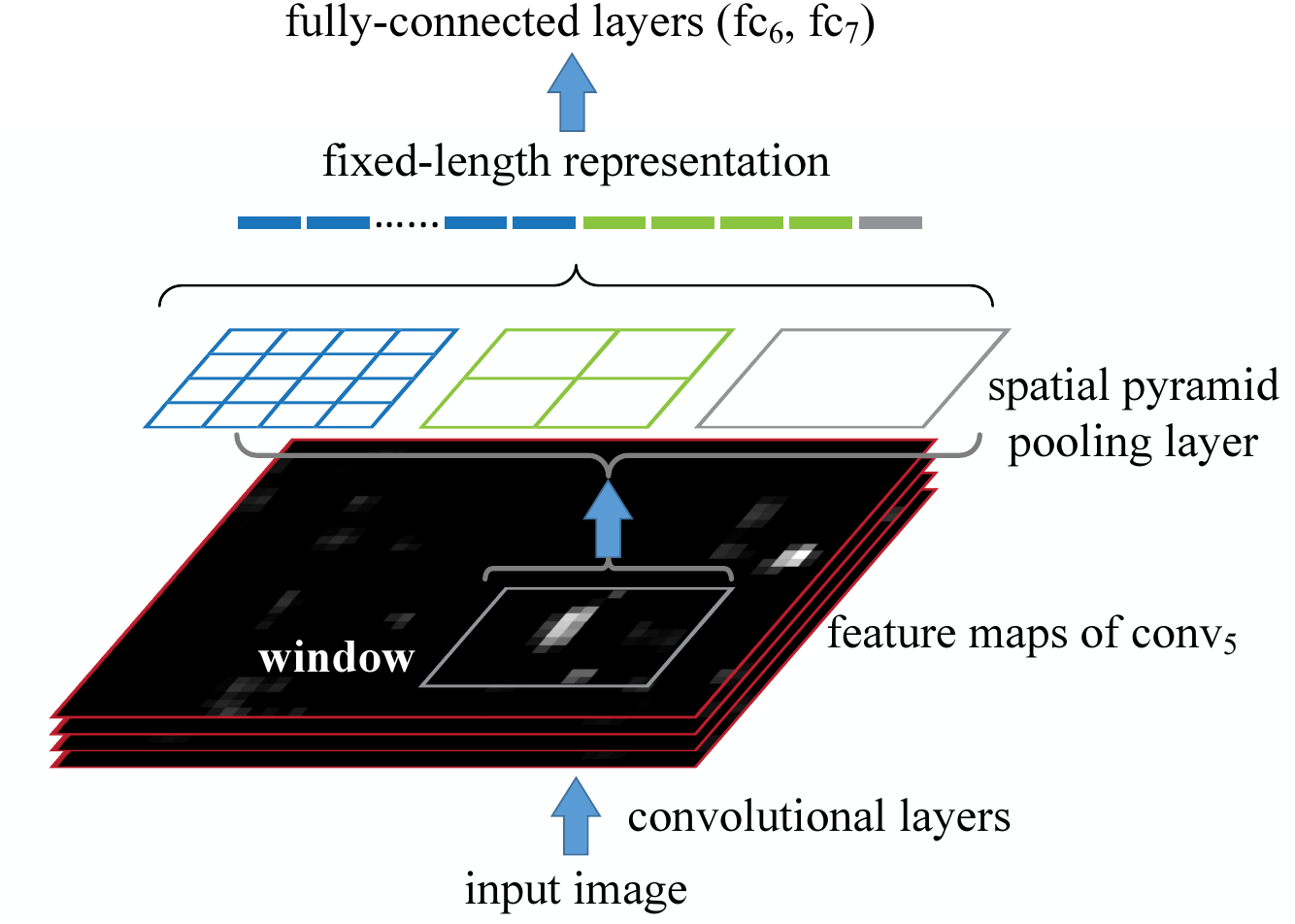}
\caption{Pooling features from arbitrary windows on feature maps. The feature maps are computed from the entire image. The pooling is performed in candidate windows.}
\label{fig:spm_det}
\end{figure}

Our SPP-net can also be used for object detection. We extract the feature maps from the entire image only once (possibly at multiple scales). Then we apply the spatial pyramid pooling on each candidate window of the feature maps to pool a fixed-length representation of this window (see Figure~\ref{fig:spm_det}). Because the time-consuming convolutions are only applied once, our method can run \emph{orders of magnitude} faster.

Our method extracts window-wise features from regions of the feature maps, while R-CNN extracts directly from image regions. In previous works, the Deformable Part Model (DPM) \cite{Felzenszwalb2010} extracts features from windows in HOG \cite{Dalal2005} feature maps, and the Selective Search (SS) method \cite{Sande2011} extracts from windows in encoded SIFT feature maps. The Overfeat detection method \cite{Sermanet2013} also extracts from windows of deep convolutional feature maps, but needs to pre-define the window size. On the contrary, our method enables feature extraction in arbitrary windows from the deep convolutional feature maps.

\subsection{Detection Algorithm}

We use the ``fast'' mode of selective search \cite{Sande2011} to generate about 2,000 candidate windows per image. Then we resize the image such that $\min(w,h)=s$, and extract the feature maps from the entire image. We use the SPP-net model of ZF-5 (single-size trained) for the time being. In each candidate window, we use a 4-level spatial pyramid (1$\times$1, 2$\times$2, 3$\times$3, 6$\times$6, totally 50 bins) to pool the features. This generates a 12,800-d (256$\times$50) representation for each window. These representations are provided to the fully-connected layers of the network.
Then we train a binary linear SVM classifier for each category on these features.

Our implementation of the SVM training follows \cite{Sande2011,Girshick2014}. We use the ground-truth windows to generate the positive samples. The negative samples are those overlapping a positive window by at most 30\% (measured by the intersection-over-union (IoU) ratio). Any negative sample is removed if it overlaps another negative sample by more than 70\%. We apply the standard hard negative mining \cite{Felzenszwalb2010} to train the SVM. This step is iterated once. It takes less than 1 hour to train SVMs for all 20 categories.
In testing, the classifier is used to score the candidate windows. Then we use non-maximum suppression \cite{Felzenszwalb2010} (threshold of 30\%) on the scored windows.

Our method can be improved by multi-scale feature extraction. We resize the image such that $\min(w,h)=s\in S = \{480, 576, 688, 864, 1200\}$, and compute the feature maps of conv$_5$ for each scale. One strategy of combining the features from these scales is to pool them channel-by-channel. But we empirically find that another strategy provides better results. For each candidate window, we choose a single scale $s\in S$ such that the scaled candidate window has a number of pixels closest to 224$\times$224. Then we only use the feature maps extracted from this scale to compute the feature of this window. If the pre-defined scales are dense enough and the window is approximately square, our method is roughly equivalent to resizing the window to 224$\times$224 and then extracting features from it. Nevertheless, our method only requires computing the feature maps once (at each scale) from the entire image, regardless of the number of candidate windows.

We also fine-tune our pre-trained network, following \cite{Girshick2014}. Since our features are pooled from the conv$_5$ feature maps from windows of any sizes, for simplicity we only fine-tune the fully-connected layers. In this case, the data layer accepts the fixed-length pooled features after conv$_5$, and the fc$_{6,7}$ layers and a new 21-way (one extra negative category) fc$_8$ layer follow. The fc$_8$ weights are initialized with a Gaussian distribution of $\sigma$=0.01. We fix all the learning rates to 1e-4 and then adjust to 1e-5 for all three layers. During fine-tuning, the positive samples are those overlapping with a ground-truth window by $[0.5, 1]$, and the negative samples by $[0.1, 0.5)$. In each mini-batch, 25\% of the samples are positive. We train 250k mini-batches using the learning rate 1e-4, and then 50k mini-batches using 1e-5. Because we only fine-tune the fc layers, the training is very fast and takes about 2 hours on the GPU (excluding pre-caching feature maps which takes about 1 hour).
Also following \cite{Girshick2014}, we use bounding box regression to post-process the prediction windows. The features used for regression are the pooled features from conv$_5$ (as a counterpart of the pool$_5$ features used in \cite{Girshick2014}). The windows used for the regression training are those overlapping with a ground-truth window by at least 50\%.

\setlength{\tabcolsep}{5pt}
\begin{table}[t]
\small
\begin{center}
\begin{tabular}{c|c c c}
\hline
 & SPP (1-sc) & SPP (5-sc) & R-CNN\\
 & \footnotesize{(ZF-5)} & \footnotesize{(ZF-5)} & \footnotesize{(Alex-5)}\\
\hline
pool$_5$ &              43.0 & \underline{44.9} & 44.2 \\
fc$_6$   &              42.5 & 44.8 & \underline{46.2} \\
ftfc$_6$ &              52.3 & \underline{53.7} & 53.1 \\
ftfc$_7$ &              54.5 & \underline{55.2} & 54.2\\
ftfc$_7$ bb &         58.0 & \textbf{59.2} & 58.5 \\
\hline
conv time (GPU) & 0.053s & 0.293s & 8.96s\\
fc time (GPU) & 0.089s & 0.089s & 0.07s\\
\hline
total time (GPU) & 0.142s & 0.382s & 9.03s\\
speedup (\vs RCNN) & \textbf{64$\times$} & \textbf{24$\times$} & -\\
\hline
\end{tabular}
\end{center}
\caption{Detection results (mAP) on Pascal VOC 2007. ``ft'' and ``bb'' denote fine-tuning and  bounding box regression.}
\label{tab:detection_layers}
~\\
\small
\begin{center}
\begin{tabular}{c|c c c}
\hline
 & SPP (1-sc) & SPP (5-sc) & R-CNN\\
 & \footnotesize{(ZF-5)} & \footnotesize{(ZF-5)} & \footnotesize{(ZF-5)}\\
\hline
ftfc$_7$ &              54.5 & \underline{55.2} & 55.1\\
ftfc$_7$ bb &         58.0 & \textbf{59.2} & \textbf{59.2} \\
\hline
conv time (GPU) & 0.053s & 0.293s & 14.37s\\
fc time (GPU) & 0.089s & 0.089s & 0.089s\\
\hline
total time (GPU) & 0.142s & 0.382s & 14.46s\\
speedup (\vs RCNN) & \textbf{102$\times$} & \textbf{38$\times$} & -\\
\hline
\end{tabular}
\end{center}
\caption{Detection results (mAP) on Pascal VOC 2007, \textbf{using the same pre-trained model} of SPP (ZF-5).}
\label{tab:detection_layers_zf}
\end{table}

\subsection{Detection Results}

We evaluate our method on the detection task of the Pascal VOC 2007 dataset.
Table~\ref{tab:detection_layers} shows our results on various layers, by using 1-scale ($s$=688) or 5-scale.
Here the R-CNN results are as reported in \cite{Girshick2014} using the AlexNet \cite{Krizhevsky2012} with 5 conv layers.
Using the pool$_5$ layers (in our case the pooled features), our result (44.9\%) is comparable with R-CNN's result (44.2\%).
But using the non-fine-tuned fc$_6$ layers, our results are inferior. An explanation is that our fc layers are pre-trained using image regions, while in the detection case they are used on the feature map regions. The feature map regions can have strong activations near the window boundaries, while the image regions may not. This difference of usages can be addressed by fine-tuning. Using the fine-tuned fc layers (ftfc$_{6,7}$), our results are comparable with or slightly better than the fine-tuned results of R-CNN. After bounding box regression, our 5-scale result (\textbf{59.2}\%) is 0.7\% better than R-CNN (58.5\%), and our 1-scale result (58.0\%) is 0.5\% worse.

In Table~\ref{tab:detection_layers_zf} we further compare with R-CNN using the same pre-trained model of SPPnet (ZF-5). In this case, our method and R-CNN have comparable averaged scores. The R-CNN result is boosted by this pre-trained model. This is because of the better architecture of ZF-5 than AlexNet, and also because of the multi-level pooling of SPPnet (if using the no-SPP ZF-5, the R-CNN result drops). Table~\ref{tab:detection_all} shows the results for each category.

Table~\ref{tab:detection_all} also includes additional methods.
Selective Search (SS) \cite{Sande2011} applies spatial pyramid matching on SIFT feature maps. DPM \cite{Felzenszwalb2010} and Regionlet \cite{Wang2013} are based on HOG features \cite{Dalal2005}. The Regionlet method improves to 46.1\% \cite{Will2014} by combining various features including conv$_5$.
DetectorNet \cite{Szegedy2013} trains a deep network that outputs pixel-wise object masks. This method only needs to apply the deep network once to the entire image, as is the case for our method. But this method has lower mAP (30.5\%).

\setlength{\tabcolsep}{2.1pt}
\begin{table*}[t]
\begin{center}
\begin{tabularx}{\textwidth}{y|x|xxxxxxxxxxxxxxxxxxxx}
  \hline
  method & \ct{mAP} & \ct{areo} & \ct{bike} & \ct{bird} & \ct{boat} & \ct{bottle} & \ct{bus} & \ct{car} & \ct{cat} & \ct{chair} & \ct{cow} & \ct{table} & \ct{dog} & \ct{horse} & \ct{mbike} & \ct{person} & \ct{plant} & \ct{sheep} & \ct{sofa} & \ct{train} & \ct{tv}\\
  \hline
  DPM \cite{Felzenszwalb2010} & 33.7 & 33.2 & 60.3	& 10.2 & 16.1 & 27.3 & 54.3 & 58.2 & 23.0 & 20.0 & 24.1 & 26.7 & 12.7 & 58.1 & 48.2 & 43.2 & 12.0 & 21.1 & 36.1 & 46.0 & 43.5\\
  SS \cite{Sande2011} & 33.8 & 43.5 & 46.5 & 10.4 & 12.0 & 9.3 & 49.4 & 53.7 & 39.4 & 12.5 & 36.9 & 42.2 & 26.4 & 47.0 & 52.4 & 23.5 & 12.1 & 29.9 & 36.3 & 42.2 & 48.8\\
  Regionlet \cite{Wang2013} & 41.7 & 54.2 & 52.0 & 20.3 & 24.0 & 20.1 & 55.5 & 68.7 & 42.6 & 19.2 & 44.2 & 49.1 & 26.6 & 57.0 & 54.5 & 43.4 & 16.4 & 36.6 & 37.7 & 59.4 & 52.3\\
  DetNet \cite{Szegedy2013} & 30.5 & 29.2 & 35.2 & 19.4 & 16.7 & 3.7 & 53.2 & 50.2 & 27.2 & 10.2 & 34.8 & 30.2 & 28.2 & 46.6 & 41.7 & 26.2 & 10.3 & 32.8 & 26.8 & 39.8 & 47.0\\
  \hline
  RCNN ftfc$_7$ (A5) & 54.2 & 64.2 & 69.7 & 50.0 & 41.9 & 32.0 & 62.6 & 71.0 & 60.7 & 32.7 & 58.5 & 46.5 & 56.1 & 60.6 & 66.8 & 54.2 & 31.5 & 52.8 & 48.9 & 57.9 & 64.7\\
  RCNN ftfc$_7$ (ZF5) & 55.1 & 64.8 & 68.4 & 47.0 & 39.5 & 30.9 & 59.8 & 70.5 & 65.3 & 33.5 & 62.5 & 50.3 & 59.5 & 61.6 & 67.9	& 54.1 & 33.4 & 57.3 & 52.9 & 60.2 & 62.9\\
  SPP ftfc$_7$ (ZF5) & 55.2  & 65.5  & 65.9  & 51.7  & 38.4  & 32.7  & 62.6  & 68.6  & 69.7  & 33.1  & 66.6  & 53.1  & 58.2  & 63.6  & 68.8  & 50.4  & 27.4  & 53.7  & 48.2  & 61.7  & 64.7 \\
  \hline
  RCNN bb (A5) & 58.5  & 68.1  & 72.8  & 56.8  & \cb{43.0}  & 36.8  & \cb{66.3}  & 74.2  & 67.6  & 34.4  & 63.5  & 54.5 & 61.2  & 69.1  & 68.6  & 58.7  & 33.4  & 62.9  & 51.1  & 62.5  & 64.8\\
  RCNN bb (ZF5) & \cb{59.2} & 68.4 & \cb{74.0} & 54.0 & 40.9 & 35.2 & 64.1 & \cb{74.4} & 69.8 & \cb{35.5} & 66.9 & 53.8 & \cb{64.2} & 69.9 & 69.6 & \cb{58.9} & \cb{36.8} & \cb{63.4} & \cb{56.0} & 62.8 & 64.9\\
  SPP bb (ZF5) & \cb{59.2}  & \cb{68.6}  & 69.7  & \cb{57.1}  & 41.2  & \cb{40.5}  & \cb{66.3}  & 71.3  & \cb{72.5}  & 34.4  & \cb{67.3}  & \cb{61.7}  & 63.1  & \cb{71.0}  & \cb{69.8}  & 57.6  & 29.7  & 59.0  & 50.2  & \cb{65.2}  & \cb{68.0} \\
  \hline
\end{tabularx}
\end{center}
\caption{Comparisons of detection results on Pascal VOC 2007.}
\label{tab:detection_all}
\end{table*}
\begin{table*}[t]
\begin{center}
\begin{tabularx}{\textwidth}{x|x|xxxxxxxxxxxxxxxxxxxx}
  \hline
  method & mAP & \ct{areo} & \ct{bike} & \ct{bird} & \ct{boat} & \ct{bottle} & \ct{bus} & \ct{car} & \ct{cat} & \ct{chair} & \ct{cow} & \ct{table} & \ct{dog} & \ct{horse} & \ct{mbike} & \ct{person} & \ct{plant} & \ct{sheep} & \ct{sofa} & \ct{train} & \ct{tv}\\
  \hline
  SPP-net (1) & 59.2  & \cb{68.6}  & 69.7  & 57.1  & 41.2  & 40.5  & 66.3  & 71.3  & 72.5  & 34.4  & \cb{67.3}  & 61.7  & 63.1  & 71.0  & 69.8  & 57.6  & 29.7  & 59.0  & 50.2  & 65.2  & 68.0 \\
  SPP-net (2) & 59.1  & 65.7  & 71.4  & 57.4  & \cb{42.4}  & 39.9  & 67.0  & 71.4  & 70.6  & 32.4  & 66.7  & 61.7  & 64.8  & 71.7  & 70.4  & 56.5  & 30.8  & 59.9  & 53.2  & 63.9  & 64.6\\
  \hline
  ~combination~ & \cb{60.9} & 68.5 & \cb{71.7} & \cb{58.7} & 41.9 & \cb{42.5} & \cb{67.7} & \cb{72.1} & \cb{73.8} & \cb{34.7} & 67.0 & \cb{63.4} & \cb{66.0} & \cb{72.5} & \cb{71.3} & \cb{58.9} & \cb{32.8} & \cb{60.9} & \cb{56.1} & \cb{67.9} & \cb{68.8}\\
\hline
\end{tabularx}
\end{center}
\caption{Detection results on VOC 2007 using model combination. The results of both models use ``ftfc$_7$ bb''.}
\label{tab:detection_comb}
\end{table*}

\subsection{Complexity and Running Time}

Despite having comparable accuracy, our method is much faster than R-CNN. The complexity of the convolutional feature computation in R-CNN is $O(n\cdot227^2)$ with the window number $n$ ($\sim$2000). This complexity of our method is $O(r\cdot s^2)$ at a scale $s$, where $r$ is the aspect ratio. Assume $r$ is about 4/3. In the single-scale version when $s=688$, this complexity is about 1/160 of R-CNN's; in the 5-scale version, this complexity is about 1/24 of R-CNN's.

In Table~\ref{tab:detection_layers_zf}, we provide a fair comparison on the running time of the feature computation \textbf{using the same SPP (ZF-5) model}.
The implementation of R-CNN is from the code published by the authors implemented in \emph{Caffe} \cite{Jia2013}. We also implement our feature computation in \emph{Caffe}. In Table~\ref{tab:detection_layers_zf} we evaluate the average time of 100 random VOC images using GPU.
R-CNN takes 14.37s per image for convolutions, while our 1-scale version takes only 0.053s per image.
So ours is 270$\times$ faster than R-CNN.
Our 5-scale version takes 0.293s per image for convolutions, so is 49$\times$ faster than R-CNN.
Our convolutional feature computation is so fast that the computational time of fc layers takes a considerable portion. Table~\ref{tab:detection_layers_zf} shows that the GPU time of computing the 4,096-d fc$_7$ features is 0.089s per image. Considering both convolutional and fully-connected features, our 1-scale version is \textbf{102$\times$} faster than R-CNN and is 1.2\% inferior; our 5-scale version is \textbf{38$\times$} faster and has comparable results.

We also compares the running time in Table~\ref{tab:detection_layers} where R-CNN uses AlexNet \cite{Krizhevsky2012} as is in the original paper \cite{Girshick2014}. Our method is 24$\times$ to 64$\times$ faster. Note that the AlexNet \cite{Krizhevsky2012} has the same number of filters as our ZF-5 on each conv layer. The AlexNet is faster because it uses splitting on some layers, which was designed for two GPUs in \cite{Krizhevsky2012}.

We further achieve an efficient full system with the help of the recent window proposal method \cite{Zitnick2014}.
The Selective Search (SS) proposal \cite{Sande2011} takes about 1-2 seconds per image on a CPU. The method of EdgeBoxes \cite{Zitnick2014} only takes $\sim$ 0.2s. Note that it is sufficient to use a fast proposal method during testing only.
Using the same model trained as above (using SS), we test proposals generated by EdgeBoxes only. The mAP is 52.8 without bounding box regression. This is reasonable considering that EdgeBoxes are not used for training.
Then we use both SS and EdgeBox as proposals in the training stage, and adopt only EdgeBoxes in the testing stage. The mAP is 56.3 without bounding box regression, which is better than 55.2 (Table~\ref{tab:detection_layers_zf}) due to additional training samples.
In this case, the overall testing time is $\sim$0.5s per image including all steps (proposal and recognition). This makes our method practical for real-world applications.

\subsection{Model Combination for Detection}

Model combination is an important strategy for boosting CNN-based classification accuracy \cite{Krizhevsky2012}. We propose a simple combination method for detection.

We pre-train another network in ImageNet, using the same structure but different random initializations. Then we repeat the above detection algorithm. Table~\ref{tab:detection_comb} (SPP-net (2)) shows the results of this network. Its mAP is comparable with the first network (59.1\% \vs 59.2\%), and outperforms the first network in 11 categories.

Given the two models, we first use either model to score all candidate windows on the test image.
Then we perform non-maximum suppression on the union of the two sets of candidate windows (with their scores).
A more confident window given by one method can suppress those less confident given by the other method. After combination, the mAP is boosted to \textbf{60.9\%} (Table~\ref{tab:detection_comb}). In 17 out of all 20 categories the combination performs better than either individual model. This indicates that the two models are complementary.

We further find that the complementarity is mainly because of the convolutional layers. We have tried to combine two randomly initialized fine-tuned results of the same convolutional model, and found no gain.

\subsection{ILSVRC 2014 Detection}

The ILSVRC 2014 detection \cite{Russakovsky2014} task involves 200 categories. There are $\sim$450k/20k/40k images in the training/validation/testing sets. We focus on the task of the provided-data-only track (the 1000-category CLS training data is not allowed to use).

There are three major differences between the detection (DET) and classification (CLS) training datasets, which greatly impacts the pre-training quality. First, the DET training data is merely 1/3 of the CLS training data. This seems to be a fundamental challenge of the provided-data-only DET task. Second, the category number of DET is 1/5 of CLS. To overcome this problem, we harness the provided subcategory labels\footnote{Using the provided subcategory labels is allowed, as is explicitly stated in the competition introduction.} for pre-training. There are totally 499 non-overlapping subcategories (\ie, the leaf nodes in the provided category hierarchy). So we pre-train a 499-category network on the DET training set. Third, the distributions of object scales are different between DET/CLS training sets. The dominant object scale in CLS is about 0.8 of the image length, but in DET is about 0.5. To address the scale difference, we resize each training image to $\min(w,h)=400$ (instead of $256$), and randomly crop $224\times224$ views for training. A crop is only used when it overlaps with a ground truth object by at least 50\%.

We verify the effect of pre-training on Pascal VOC 2007. For a CLS-pre-training baseline, we consider the pool$_5$ features (mAP 43.0\% in Table~\ref{tab:detection_layers}). Replaced with a 200-category network pre-trained on DET, the mAP significantly drops to 32.7\%. A 499-category pre-trained network improves the result to 35.9\%. Interestingly, even if the amount of training data do not increase, training a network of more categories boosts the feature quality. Finally, training with $\min(w,h)=400$ instead of $256$ further improves the mAP to 37.8\%. Even so, we see that there is still a considerable gap to the CLS-pre-training result. This indicates the importance of big data to deep learning.

For ILSVRC 2014, we train a 499-category Overfeat-7 SPP-net. The remaining steps are similar to the VOC 2007 case. Following \cite{Girshick2014}, we use the validation set to generate the positive/negative samples, with windows proposed by the selective search fast mode. The training set only contributes positive samples using the ground truth windows. We fine-tune the fc layers and then train the SVMs using the samples in both validation and training sets. The bounding box regression is trained on the validation set.

Our single model leads to 31.84\% mAP in the ILSVRC 2014 \textbf{testing} set \cite{Russakovsky2014}. We combine six similar models using the strategy introduced in this paper. The mAP is \textbf{35.11\%} in the testing set \cite{Russakovsky2014}. This result ranks \#2 in the provided-data-only track of ILSVRC 2014 (Table~\ref{tab:ilsvrc14_DET}) \cite{Russakovsky2014}. The winning result is 37.21\% from NUS, which uses contextual information.

Our system still shows great advantages on speed for this dataset. It takes our single model 0.6 seconds (0.5 for conv, 0.1 for fc, excluding proposals) per testing image on a GPU extracting convolutional features from all 5 scales. Using the same model, it takes 32 seconds per image in the way of RCNN. For the 40k testing images, our method requires 8~GPU$\cdot$hours to compute convolutional features, while RCNN would require 15~GPU$\cdot$days.

\setlength{\tabcolsep}{8pt}
\begin{table}[t]
\footnotesize
\begin{center}
\begin{tabular}{cc|c}
\hline
rank & team & mAP\\
\hline
1 &  NUS & \textbf{37.21}\\
2 & \underline{ours} & \underline{35.11}\\
3 & UvA & 32.02\\
- & (our single-model) & (31.84)\\
4 & Southeast-CASIA & 30.47\\
5 & 1-HKUST & 28.86\\
6 & CASIA\_CRIPAC\_2 & 28.61\\
\hline
\end{tabular}
\caption{The competition results of ILSVRC 2014 detection (provided-data-only track) \cite{Russakovsky2014}. The best entry of each team is listed.}
\label{tab:ilsvrc14_DET}
\end{center}
\end{table}

\renewcommand{\baselinestretch}{0.8}
\section{Conclusion}

SPP is a flexible solution for handling different scales, sizes, and aspect ratios. These issues are important in visual recognition, but received little consideration in the context of deep networks. We have suggested a solution to train a deep network with a spatial pyramid pooling layer. The resulting SPP-net shows outstanding accuracy in classification/detection tasks and greatly accelerates DNN-based detection. Our studies also show that many time-proven techniques/insights in computer vision can still play important roles in deep-networks-based recognition.

\begin{figure*}[t]
\center
\includegraphics[width=0.85\linewidth]{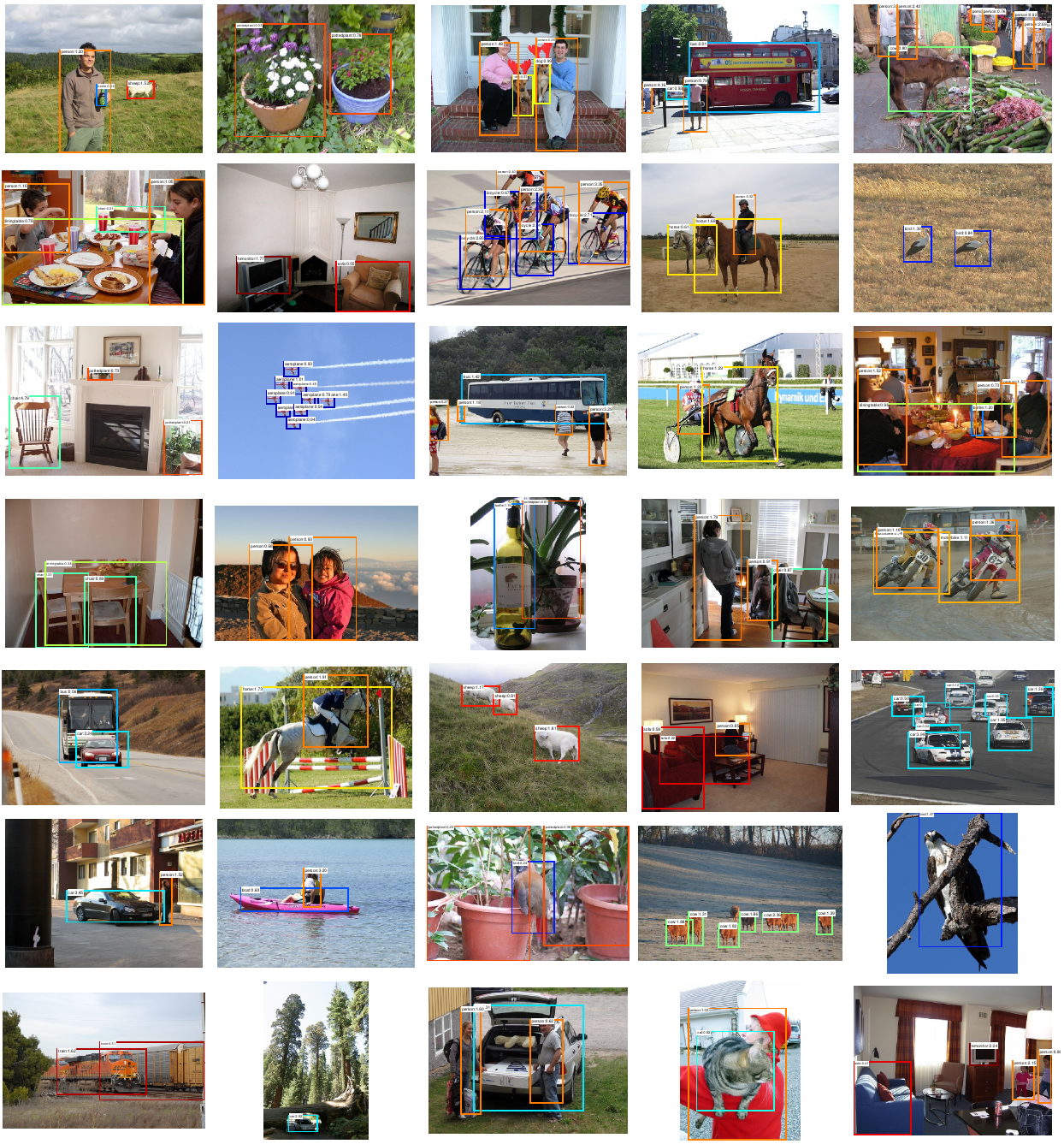}
\caption{Example detection results of ``SPP-net ftfc$_7$ bb'' on the Pascal VOC 2007 testing set (59.2\% mAP). All windows with scores $>$ 0 are shown. The predicted category/score are marked. The window color is associated with the predicted category. These images are manually selected because we find them impressive. Visit our project website to see all 4,952 detection results in the testing set.}
\label{fig:res}
\end{figure*}

\appendices
\section{}
In the appendix, we describe some implementation details:

\vspace{6pt}
\noindent\textbf{Mean Subtraction.}

The 224$\times$224 cropped training/testing images are often pre-processed by subtracting the per-pixel mean \cite{Krizhevsky2012}. When input images are in any sizes, the fixed-size mean image is not directly applicable. In the ImageNet dataset, we warp the 224$\times$224 mean image to the desired size and then subtract it.
In Pascal VOC 2007 and Caltech101, we use the constant mean (128) in all the experiments.

\vspace{6pt}
\noindent\textbf{Implementation of Pooling Bins.}

We use the following implementation to handle all bins when applying the network. Denote the width and height of the conv$_5$ feature maps (can be the full image or a window) as $w$ and $h$. For a pyramid level with $n$$\times$$n$ bins, the $(i,j)$-th bin is in the range of $[\lfloor\frac{i-1}{n}w\rfloor, \lceil\frac{i}{n}w\rceil]\times[\lfloor\frac{j-1}{n}h\rfloor, \lceil\frac{j}{n}h\rceil]$. Intuitively, if rounding is needed, we take the floor operation on the left/top boundary and ceiling on the right/bottom boundary.

\vspace{6pt}
\noindent\textbf{Mapping a Window to Feature Maps.}

In the detection algorithm (and multi-view testing on feature maps), a window is given in the image domain, and we use it to crop the convolutional feature maps (\eg, conv$_5$) which have been sub-sampled several times. So we need to align the window on the feature maps.

In our implementation, we project the corner point of a window onto a pixel in the feature maps, such that this corner point in the image domain is closest to the center of the receptive field of that feature map pixel.
The mapping is complicated by the padding of all convolutional and pooling layers. To simplify the implementation, during deployment we pad $\lfloor p/2 \rfloor$ pixels for a layer with a filter size of $p$. As such, for a response centered at $(x', y')$ , its effective receptive field in the image domain is centered at $(x,y)=(Sx', Sy')$ where $S$ is the product of all previous strides. In our models, $S=16$ for ZF-5 on conv$_5$, and $S=12$ for Overfeat-5/7 on conv$_{5/7}$. Given a window in the image domain, we project the left (top) boundary by: $x' = \lfloor x /S\rfloor+1$ and the right (bottom) boundary $x' = \lceil x /S \rceil-1$.
If the padding is not $\lfloor p/2 \rfloor$, we need to add a proper offset to $x$.

\bibliographystyle{IEEEtran}
\bibliography{IEEEabrv,sppnet_v4}

\begin{thebibliography}{10}
\providecommand{\url}[1]{#1}
\csname url@samestyle\endcsname
\providecommand{\newblock}{\relax}
\providecommand{\bibinfo}[2]{#2}
\providecommand{\BIBentrySTDinterwordspacing}{\spaceskip=0pt\relax}
\providecommand{\BIBentryALTinterwordstretchfactor}{4}
\providecommand{\BIBentryALTinterwordspacing}{\spaceskip=\fontdimen2\font plus
\BIBentryALTinterwordstretchfactor\fontdimen3\font minus
  \fontdimen4\font\relax}
\providecommand{\BIBforeignlanguage}[2]{{%
\expandafter\ifx\csname l@#1\endcsname\relax
\typeout{** WARNING: IEEEtran.bst: No hyphenation pattern has been}%
\typeout{** loaded for the language `#1'. Using the pattern for}%
\typeout{** the default language instead.}%
\else
\language=\csname l@#1\endcsname
\fi
#2}}
\providecommand{\BIBdecl}{\relax}
\BIBdecl

\bibitem{LeCun1989}
Y.~LeCun, B.~Boser, J.~S. Denker, D.~Henderson, R.~E. Howard, W.~Hubbard, and
  L.~D. Jackel, ``Backpropagation applied to handwritten zip code
  recognition,'' \emph{Neural computation}, 1989.

\bibitem{Deng2009}
J.~Deng, W.~Dong, R.~Socher, L.-J. Li, K.~Li, and L.~Fei-Fei, ``Imagenet: A
  large-scale hierarchical image database,'' in \emph{CVPR}, 2009.

\bibitem{Krizhevsky2012}
A.~Krizhevsky, I.~Sutskever, and G.~Hinton, ``Imagenet classification with deep
  convolutional neural networks,'' in \emph{NIPS}, 2012.

\bibitem{Zeiler2013}
M.~D. Zeiler and R.~Fergus, ``Visualizing and understanding convolutional
  neural networks,'' \emph{arXiv:1311.2901}, 2013.

\bibitem{Sermanet2013}
P.~Sermanet, D.~Eigen, X.~Zhang, M.~Mathieu, R.~Fergus, and Y.~LeCun,
  ``Overfeat: Integrated recognition, localization and detection using
  convolutional networks,'' \emph{arXiv:1312.6229}, 2013.

\bibitem{Chatfield2014}
A.~V. K.~Chatfield, K.~Simonyan and A.~Zisserman, ``Return of the devil in the
  details: Delving deep into convolutional nets,'' in \emph{ArXiv:1405.3531},
  2014.

\bibitem{Girshick2014}
R.~Girshick, J.~Donahue, T.~Darrell, and J.~Malik, ``Rich feature hierarchies
  for accurate object detection and semantic segmentation,'' in \emph{CVPR},
  2014.

\bibitem{Will2014}
W.~Y. Zou, X.~Wang, M.~Sun, and Y.~Lin, ``Generic object detection with dense
  neural patterns and regionlets,'' in \emph{ArXiv:1404.4316}, 2014.

\bibitem{Razavian2014}
A.~S. Razavian, H.~Azizpour, J.~Sullivan, and S.~Carlsson, ``Cnn features
  off-the-shelf: An astounding baseline for recogniton,'' in \emph{CVPR 2014,
  DeepVision Workshop}, 2014.

\bibitem{Yaniv2014}
Y.~Taigman, M.~Yang, M.~Ranzato, and L.~Wolf, ``Deepface: Closing the gap to
  human-level performance in face verification,'' in \emph{CVPR}, 2014.

\bibitem{Ning2014}
N.~Zhang, M.~Paluri, M.~Ranzato, T.~Darrell, and L.~Bourdevr, ``Panda: Pose
  aligned networks for deep attribute modeling,'' in \emph{CVPR}, 2014.

\bibitem{Yunchao2014}
Y.~Gong, L.~Wang, R.~Guo, and S.~Lazebnik, ``Multi-scale orderless pooling of
  deep convolutional activation features,'' in \emph{ArXiv:1403.1840}, 2014.

\bibitem{Donahue2013}
J.~Donahue, Y.~Jia, O.~Vinyals, J.~Hoffman, N.~Zhang, E.~Tzeng, and T.~Darrell,
  ``Decaf: A deep convolutional activation feature for generic visual
  recognition,'' \emph{arXiv:1310.1531}, 2013.

\bibitem{Grauman2005}
K.~Grauman and T.~Darrell, ``The pyramid match kernel: Discriminative
  classification with sets of image features,'' in \emph{ICCV}, 2005.

\bibitem{Lazebnik2006}
S.~Lazebnik, C.~Schmid, and J.~Ponce, ``Beyond bags of features: Spatial
  pyramid matching for recognizing natural scene categories,'' in \emph{CVPR},
  2006.

\bibitem{Sivic2003}
J.~Sivic and A.~Zisserman, ``Video google: a text retrieval approach to object
  matching in videos,'' in \emph{ICCV}, 2003.

\bibitem{Yang2009}
J.~Yang, K.~Yu, Y.~Gong, and T.~Huang, ``Linear spatial pyramid matching using
  sparse coding for image classification,'' in \emph{CVPR}, 2009.

\bibitem{Wang2010}
J.~Wang, J.~Yang, K.~Yu, F.~Lv, T.~Huang, and Y.~Gong, ``Locality-constrained
  linear coding for image classification,'' in \emph{CVPR}, 2010.

\bibitem{Perronnin2010}
F.~Perronnin, J.~S{\'a}nchez, and T.~Mensink, ``Improving the fisher kernel for
  large-scale image classification,'' in \emph{ECCV}, 2010.

\bibitem{Sande2011}
K.~E. van~de Sande, J.~R. Uijlings, T.~Gevers, and A.~W. Smeulders,
  ``Segmentation as selective search for object recognition,'' in \emph{ICCV},
  2011.

\bibitem{Fei-Fei2007}
L.~Fei-Fei, R.~Fergus, and P.~Perona, ``Learning generative visual models from
  few training examples: An incremental bayesian approach tested on 101 object
  categories,'' \emph{CVIU}, 2007.

\bibitem{Everingham2007}
M.~Everingham, L.~Van~Gool, C.~K.~I. Williams, J.~Winn, and A.~Zisserman, ``The
  {PASCAL} {V}isual {O}bject {C}lasses {C}hallenge 2007 {(VOC2007)}
  {R}esults,'' 2007.

\bibitem{Felzenszwalb2010}
P.~F. Felzenszwalb, R.~B. Girshick, D.~McAllester, and D.~Ramanan, ``Object
  detection with discriminatively trained part-based models,'' \emph{PAMI},
  2010.

\bibitem{Dalal2005}
N.~Dalal and B.~Triggs, ``Histograms of oriented gradients for human
  detection,'' in \emph{CVPR}, 2005.

\bibitem{Zitnick2014}
C.~L. Zitnick and P.~Doll{\'a}r, ``Edge boxes: Locating object proposals from
  edges,'' in \emph{ECCV}, 2014.

\bibitem{Russakovsky2014}
O.~Russakovsky, J.~Deng, H.~Su, J.~Krause, S.~Satheesh, S.~Ma, Z.~Huang,
  A.~Karpathy, A.~Khosla, M.~Bernstein \emph{et~al.}, ``Imagenet large scale
  visual recognition challenge,'' \emph{arXiv:1409.0575}, 2014.

\bibitem{Chatfield2011}
K.~Chatfield, V.~Lempitsky, A.~Vedaldi, and A.~Zisserman, ``The devil is in the
  details: an evaluation of recent feature encoding methods,'' in \emph{BMVC},
  2011.

\bibitem{Coates2011}
A.~Coates and A.~Ng, ``The importance of encoding versus training with sparse
  coding and vector quantization,'' in \emph{ICML}, 2011.

\bibitem{Lowe2004}
D.~G. Lowe, ``Distinctive image features from scale-invariant keypoints,''
  \emph{IJCV}, 2004.

\bibitem{Gemert2008}
J.~C. van Gemert, J.-M. Geusebroek, C.~J. Veenman, and A.~W. Smeulders,
  ``Kernel codebooks for scene categorization,'' in \emph{ECCV}, 2008.

\bibitem{Lin2013}
M.~Lin, Q.~Chen, and S.~Yan, ``Network in network,'' \emph{arXiv:1312.4400},
  2013.

\bibitem{Szegedy2014}
C.~Szegedy, W.~Liu, Y.~Jia, P.~Sermanet, S.~Reed, D.~Anguelov, D.~Erhan,
  V.~Vanhoucke, and A.~Rabinovich, ``Going deeper with convolutions,''
  \emph{arXiv:1409.4842}, 2014.

\bibitem{Simonyan2014}
K.~Simonyan and A.~Zisserman, ``Very deep convolutional networks for
  large-scale image recognition,'' \emph{arXiv:1409.1556}, 2014.

\bibitem{Oquab2014}
M.~Oquab, L.~Bottou, I.~Laptev, J.~Sivic \emph{et~al.}, ``Learning and
  transferring mid-level image representations using convolutional neural
  networks,'' in \emph{CVPR}, 2014.

\bibitem{Jia2013}
Y.~Jia, ``{Caffe}: An open source convolutional architecture for fast feature
  embedding,'' \url{http://caffe.berkeleyvision.org/}, 2013.

\bibitem{Howard2013}
A.~G. Howard, ``Some improvements on deep convolutional neural network based
  image classification,'' \emph{ArXiv:1312.5402}, 2013.

\bibitem{Jegou2012}
H.~Jegou, F.~Perronnin, M.~Douze, J.~Sanchez, P.~Perez, and C.~Schmid,
  ``Aggregating local image descriptors into compact codes,'' \emph{TPAMI},
  vol.~34, no.~9, pp. 1704--1716, 2012.

\bibitem{Chang2011}
C.-C. Chang and C.-J. Lin, ``Libsvm: a library for support vector machines,''
  \emph{ACM Transactions on Intelligent Systems and Technology (TIST)}, 2011.

\bibitem{Wang2013}
X.~Wang, M.~Yang, S.~Zhu, and Y.~Lin, ``Regionlets for generic object
  detection,'' in \emph{ICCV}, 2013.

\bibitem{Szegedy2013}
C.~Szegedy, A.~Toshev, and D.~Erhan, ``Deep neural networks for object
  detection,'' in \emph{NIPS}, 2013.

\end{thebibliography}

\section*{Changelog}

\noindent\textbf{arXiv v1}. Initial technical report for ECCV 2014 paper.

\vspace{6pt}
\noindent\textbf{arXiv v2}. Submitted version for TPAMI. Includes extra experiments of SPP on various architectures. Includes details for ILSVRC 2014.

\vspace{6pt}
\noindent\textbf{arXiv v3}. Accepted version for TPAMI. Includes comparisons with R-CNN using the same architecture. Includes detection experiments using EdgeBoxes.

\vspace{6pt}
\noindent\textbf{arXiv v4}. Revised ``Mapping a Window to Feature Maps'' in Appendix for easier implementation.

\vfill

\end{document}